\documentclass[10pt]{article}
\usepackage[utf8]{inputenc}
\usepackage{ragged2e}
\usepackage{tablefootnote}
\usepackage[a4paper,vmargin={20mm,20mm},hmargin={20mm,20mm}]{geometry}
\usepackage{multicol}
\usepackage{multirow}
\usepackage{graphicx}
\usepackage{svg}
\usepackage{subcaption}
\usepackage{wrapfig}
\usepackage{booktabs}
\usepackage [english]{babel}

\usepackage[frozencache ,cachedir=.]{minted}

\usepackage [autostyle, english = american]{csquotes}
\MakeOuterQuote{"}

\usepackage{listings}

\usepackage[page]{appendix} 

\usepackage{color, colortbl}
\definecolor{LightGrey}{rgb}{0.9,0.9,0.9}

\usepackage{amssymb}
\usepackage{pifont}
\usepackage{adjustbox}
\newcommand{\cmark}{\ding{51}}%
\newcommand{\xmark}{\ding{55}}%

\usepackage{xcolor}

\usepackage[backend=biber,style=numeric,sorting=none,sortcites=true]{biblatex}

\usepackage[hidelinks]{hyperref}
\hypersetup{
  colorlinks=true,
  urlcolor=blue,
  linkcolor=green,
  citecolor=red
}

\title{RE-GAINS \& EnChAnT: Intelligent Tool Manipulation Systems For Enhanced Query Responses}

\author{
  Sahil Girhepuje \and
  Siva Sankar Sajeev \and
  Purvam Jain  \and
  Arya Sikder \and
  Adithya Rama Varma \and
  Ryan George \and
  Akshay Govind Srinivasan \and
  Mahendra Kurup \and
  Ashmit Sinha  \and  
  Sudip Mondal  \\ 
  \textit{Indian Institute of Technology Madras, Chennai-600036}
}

\begin{document}

\maketitle


\setlength\columnsep{14pt}
\begin{multicols}{2}


\begin{abstract}
    Large Language Models (LLMs) currently struggle with tool invocation and chaining, as they often hallucinate or miss essential steps in a sequence. We propose RE-GAINS and EnChAnT, two novel frameworks that empower LLMs to tackle complex user queries by making API calls to external tools based on tool descriptions and argument lists. Tools are chained based on the expected output, without receiving the actual results from each individual call. EnChAnT, an open-source solution, leverages an LLM format enforcer, OpenChat 3.5 (an LLM), and ToolBench's API Retriever. RE-GAINS utilizes OpenAI models and embeddings with a specialized prompt based on the $\underline{R}$easoning vi$\underline{a}$ $\underline{P}$lanning $(RAP)$ framework. Both frameworks are low cost (0.01\$ per query). Our key contribution is enabling LLMs for tool invocation and chaining using modifiable, externally described tools.
\end{abstract}

\section{Introduction}
Large Language Models (LLMs) demonstrate exceptional capabilities in handling language-based tasks. However, for LLMs to progress towards Artificial General Intelligence (AGI), they must also efficiently perform logical and mathematical operations, an area where they currently struggle \cite{huang2023reasoning}. Along similar lines, tool-augmented LLMs gain importance. This involves using LLMs to access external APIs for assistance with logical and mathematical challenges.

In our work, we have established two distinct tool-based pipelines. The first pipeline is designed to maximize efficiency, while the second is optimized for performance. Both systems are developed with specific objectives in mind. They are crafted to analyze the input query, select the appropriate tools, define the tool arguments, and sequence the deployment of these tools. Moreover, our aim is to create systems that are highly scalable and capable of processing queries from various domains. These systems are planned to maintain low operational costs and minimize the carbon footprint.

Furthermore, we have developed a novel method for high-quality data generation. This is particularly useful for fine-tuning a tool-augmented LLM. We fine-tuned multiple publicly available models like OpenChat and GPT 3.5. 


\section{Literature Review}

The literature review discussion is partitioned into five distinct subsections. We start with a generic overview of the prompting techniques of LLMs. We then examine existing literature on Tool-based LLMs. Following this, we explore benchmark datasets in our domain. We move on to check models where prevailing tools prove inadequate. Lastly, the review encompasses utilising embeddings for tool retrieval, culminating in a small experiment designed to gauge its applicability to our objectives.

\subsection{Prompting Techniques}
This section offers insights that will be useful in shaping the design and optimisation of our models. After reading through many publications and surveys\cite{Prompting-survey} on various prompting strategies, we have devised implementations to enhance the performance and capabilities of the models we have developed.


\medskip\textit{\textbf{Chain-of-Thought}} (CoT) prompting\cite{wei2023chainofthought}  can dramatically improve the multi-step reasoning abilities of vanilla LLMs. CoT explicitly encourages the LLM to generate intermediate rationales for solving a problem by providing reasoning steps in the demonstrations. Despite its success, there still needs to be more understanding of what makes CoT prompting effective \cite{wang-etal-2023-cot}. For example, employing chain-of-thought prompting with just eight exemplars on a PaLM 540B \cite{huang2022large-palm-benchmark} achieves state-of-the-art accuracy on the GSM8K benchmark for math word problems, surpassing even a fine-tuned GPT-3 with a verifier \cite{GSM8K}. However, CoT encounters challenges due to token limits. 

\medskip The limitation of CoT lies in its reliance on the model's articulated reasoning accurately reflecting its underlying thought process. CoT continuously relies on the previous thought in the line; hence, any bias gets carried through the entire thought process. This challenge is mitigated by task decomposition, where questions are broken down into independent and separate sub-questions, enhancing the model's faithfulness, also termed as \textit{\textbf{factored decomposition}} \cite{radhakrishnan2023questiondecompositionfaithfullness}. This employs multiple contexts to independently answer subquestions before consolidating the subanswers into a final response. It reduces biased reasoning and diminishes the likelihood of overlooking relevant reasoning, explicitly outlining the relationship between subquestion answers and subsequent follow-up subquestions.

\medskip \textit{\textbf{Least-to-Most prompting}} \cite{zhou2023leasttomost} involves breaking down complex problems into simpler subproblems and solving them sequentially. The approach outperforms CoT, especially when problems have more steps. It also performs well when we require generalisation to solve challenges beyond demonstration examples. GPT-3 code-davinci-002 with least-to-most prompting achieves over 99\%  accuracy on the compositional generalisation benchmark SCAN, compared to 16\%  accuracy with CoT. Similarly, compared to CoT, the GSM8K benchmark also sees a 15\% increase in one-shot accuracy for problems requiring five or more steps. Inspired by this, the idea of exploring question decomposition arises. \textit{\textbf{Plan-and-Solve prompting}} \cite{wang2023planandsolve} replaces CoT's \textit{"let's think step by step"} with \textit{"Let's first understand the problem and devise a plan to solve it. Then, let's carry out the plan and solve the problem step by step."} This approach, included in the core library of \textbf{LangChain} \cite{langchain} as `Plan-and-Execute,' outperforms zero-shot and few-shot CoT. We explored other techniques like Successive prompting\cite{Successive-prompt} and Selection Inference\cite{Selection-inference}, which do not seem helpful for our problem statement.


\medskip\textit{\textbf{Step-Back prompting}} \cite{zheng2023stepbackprompting} leverages the idea of breaking down complex problems into smaller abstractions or stepback questions. The stepback answer is used for the final solution. This approach uses a single language model in an iterative process. In conjunction with RAG \cite{lewis2021retrievalaugmentedgeneration-rag}, Step-Back Prompting may give comparable results to other powerful models like GPT. 

 \medskip\textit{\textbf{Analogical Prompting}} \cite{yasunaga2023analogical} requests a language model to recall relevant examples from the past for each query, enhancing contextual awareness. This ensures the model uses higher "level" training exemplars instead of the most primitive ones. This allows it to perform much more complex tasks.
 The method outperforms retrieved CoT with larger-scale language models like text-davinci-003. However, retrieved CoT performs better for smaller models, indicating the need for contextual exemplars. Analogical Prompting is particularly effective for tasks requiring fluid and creative thinking, showcasing applications in code completion and math tasks. The success of Analogical Prompting depends on the model's learning effectiveness while training on a good dataset.

In our review of prompting techniques for enhancing LLMs capabilities, we commence with the Chain of Thought \cite{wei2023chainofthought} method, which provides a foundation for complex reasoning. Task decomposition emerges as a critical component, with strategies like factored \cite{radhakrishnan2023questiondecompositionfaithfullness} and least-to-most prompting \cite{zhou2023leasttomost} dissecting intricate tasks into simpler subtasks. The technique of step-back prompting \cite{zheng2023stepbackprompting} further refines this approach. By asking reflective questions such as "Do I know who the user is?", models can pinpoint and retrieve missing information, which is crucial for tool manipulation. Additionally, the efficacy of analogical prompting\cite{yasunaga2023analogical} cannot be understated; by leveraging apt analogies from extensive databases, models gain significant contextual support. These integrated techniques highlight the essential role of systematic task decomposition and contextual understanding in elevating LLMs' proficiency in tool manipulation.

We have evaluated various prompting techniques and presented data in the Appendix \ref{Section: Eval Prompting Tech}, Table \ref{table: prompting eval-jsjs} and Table \ref{table: prompting eval-tsjs}. 

We now shift our attention to tool-augmented models.

\subsection{Existing literature on Tool-based LLMs}

Several noteworthy methodologies have emerged in incorporating tool-using capabilities into Large Language Models (LLMs). 

\medskip The authors of\textbf{\textit{ReAct}} \cite{yao2023react} present a unique method which employs the pre-trained model `text-davinci-002' to sequentially integrate reasoning, action, and observation by the usage of structured prompts in directing the model towards desired outcomes. The authors note that the baselines, which focus exclusively on reasoning, lead to misinformation or action, which may lack necessary reasoning capabilities. ReAct demonstrates its proficiency in creating accurate and comprehensible narratives by incorporating information from external environments, thus harmonising the reasoning-action dynamic. We believe that implementing ReAct could reduce the hallucination rate by ensuring the existence of tools that form the LLMs' thought process during each stage of its evolution.

\medskip The authors of \textbf{\textit{Toolformer}} \cite{schick2023toolformer} introduce a new LLM that is trained on a LLM annotated pre-training dataset. It exhibits prowess in solving complex problems by leveraging external APIs. While capable of recognising and determining tool usage, Toolformer is constrained by two problems: (a) \textit{A fixed set of available tools}, as new pre-training datasets need to be generated for added tools, and (b) \textit{the inability to use tools in a chain}, as API calls for each tool are generated independently. Additionally, it implements a novel self-supervised augmentation in the training dataset, leading to self-supervised training. 

\medskip The \textit{\textbf{ART}} framework, as presented by Paranjape et al. \cite{paranjape2023art}, stands out by using frozen LLMs to generate automatic multi-step decompositions for new tasks automatically. It achieves this by selecting decompositions from related functions in the task library and utilising tools from the tool library during LLM generation. Importantly, human intervention is optional, allowing for the enhancement of performance through decomposition editing. Another approach, \textit{\textbf{ChatCoT}} \cite{chen2023chatcot}, addresses tool use planning by considering tools manipulation by LLMs as the interaction between LLMs and tools, modeling it as a multi-turn conversation.

\begin{table*}[]
    \centering
    \resizebox{1.95\columnwidth}{!}{
        \begin{tabular}{lccccc}
        \toprule
        Method & API-Bank \cite{li2023apibank} &ToolBench \cite{xu2023toolbench} & APIBench \cite{patil2023gorilla} &  ToolAlpaca \cite{tang2023toolalpaca} & Our Agent \\
        \midrule
        Real-world API? & \textcolor{green}{\checkmark} & \textcolor{green}{\cmark} & \textcolor{red}{\xmark} &   \textcolor{red}{\xmark} &  \textcolor{green}{\cmark} \\
        Real API Response? & \textcolor{green}{\checkmark} &  \textcolor{green}{\cmark} & \textcolor{red}{\xmark} &  \textcolor{red}{\xmark}  & \textcolor{green}{\cmark} \\
        \rowcolor{LightGrey}
        Multi-tool Scenario? &   \textcolor{red}{\xmark}  & \textcolor{green}{\cmark} & \textcolor{red}{\xmark} & \textcolor{red}{\xmark} & \textcolor{green}{\cmark} \\ 
        API Retrieval? & \textcolor{red}{\xmark} & \textcolor{green}{\cmark} & \textcolor{green}{\cmark} &   \textcolor{red}{\xmark} & \textcolor{green}{\cmark} \\ 
        Multi-step Reasoning? & \textcolor{green}{\cmark} &  \textcolor{green}{\checkmark} & \textcolor{red}{\xmark} & \textcolor{green}{\cmark} & \textcolor{green}{\cmark} \\ \hline
        Number of tools & 53 &  3451 & 3 & 400  & $\leq$ 50\\
        Number of Real API Calls & 568  & 37204 & 0  & 0 & 0 \\ \hline
        \bottomrule
        \end{tabular}
    }
    \caption{Comparison of multiple resources from the literature on Methods related to Tool-based LLMs as given in \cite{xu2023toolbench}. The rightmost column indicates the requirements for this problem statement}
    \label{table: compare models apis}
\end{table*}

\medskip The \textbf{\textit{Chameleon}} framework \cite{lu2023chameleon} stands out as a plug-and-play compositional reasoning framework that enhances LLMs with various tools. It supports tools like search engines and Python functions. However, custom tools cannot be added. Chameleon excels in inferring the appropriate sequence of tools to compose and execute to generate a response. It is suggested that when employing GPT-4 as an exhibit, Chameleon demonstrates more consistent and rational tool selection, inferring potential constraints given in instructions.

\medskip Moving on, the \textbf{\textit{GEAR}} framework \cite{lu2023gear} emphasises the importance of a retriever in suppressing hallucination. \textit{GEAR} employs prompt rewriting through another LLM and RL-based learning for auto prompt generation at each user prompt. Hence, it focuses on chat LLMs for API documentation. The authors note that GEAR-augmented GPT-J and GPT-3 outperform counterpart tool-augmented baselines due to superior tool use. It facilitates easy working through a three-step process: fetching the best tools, ranking them, and selecting them. However, it permits only one tool per chat, akin to AutoGPT \cite{autogpt}.

\medskip Furthermore, \textbf{\textit{TALM}} \cite{parisi2022talm} augments LLMs like T5 \cite{raffel2023t5} with tools via a text-to-text API. Notably, TALM can generalise input text that is out-of-distribution to the model's training data yet solvable with access to tools. It introduces a new pipeline for fine-tuning/ creating a tool use set, a special case of a policy-gradient RL algorithm, where the LM is the policy network and is trained by a policy gradient with a binary reward
signal. It excels in knowledge-heavy question-answering tasks, showcasing its adaptability when replacing the BM-25 Wiki retriever with a public search engine. Noteworthy in this landscape is \textbf{\textit{HuggingGPT}} \cite{shen2023hugginggpt}, an LLM that leverages the Hugging Face API to solve AI tasks.

\medskip Program-Aided Language models \textbf{\textit{PAL}} \cite{gao2023pal} introduces a novel approach addressing the tendency of LLMs to make logical and arithmetic mistakes during the solution phase. PAL utilises the LLM to generate programs as intermediate reasoning steps, offloading the solution to a Python interpreter. With PAL, decomposing the natural language problem into runnable steps remains the only learning task for the LLM. Notably, PAL surpasses chain-of-thought models, providing a promising avenue for LLMs to interact with runtime environments. However, it requires improvement in handling descriptive variable names and might struggle with domain-specific knowledge. Generating examples for API documentation could enhance its performance.

\medskip Stanford's \textbf{\textit{ToolAlpaca}} \cite{tang2023toolalpaca} is designed to impart generalised tool-use abilities to compact language models with minimal human supervision. However, the model has to be trained after every new tool addition.

\medskip \textbf{\textit{Gorilla}} \cite{patil2023gorilla} stands out as a pivotal paper. It uses the LLAMA-7B model to extract correct APIs from TensorHub, HuggingFace and TorchHub. Gorilla significantly outperforms GPT-4 regarding API functionality accuracy and reducing hallucination errors. The model primarily emphasises enhancing LLMs' capability to effectively utilise various tools, prioritising practical utility over refining conversational skills. The comprehensive evaluation of Gorilla includes an extensive dataset of 11,000 API pairs. The retrieval aspect is commendable as well. However, Gorilla does present some limitations. Its reliance on a machine learning dataset could limit its applicability to other domains. The necessity for fine-tuning raises questions about the model's generalizability to diverse scenarios and its adaptability to custom APIs that it has yet to encounter during training.

\medskip \textbf{\textit{ToolBench/ToolLLM}} \cite{xu2023toolbench} showcases improvements over Gorilla as shown in Table \ref{table: compare models apis}. Notably, the model excels in assembling APIs in the correct order, providing a systematic approach to tool utilisation. One significant innovation lies in the concept of Instruction Generation, where ChatGPT is prompted to generate diverse instructions for single-tool and multi-tool scenarios sampled from various APIs. By observing how even the most sophisticated GPT-4 achieves a low pass rate for complex human instructions, they develop a novel depth-first search-based decision tree (DFSDT) to broaden the search space of LLMs and hence improve the general decision-making capability leading to a remarkable out-of-distribution (OOD) generalisation performance.

\medskip Comparatively, ToolBench/ToolLLM surpasses Gorilla on several fronts. Firstly, Gorilla's limited engagement with real-world APIs, focusing on a narrow scope with poor diversity, contrasts with ToolBench's broader approach. Secondly, while Gorilla is confined to single-tool scenarios, ToolBench recognises the real-world necessity of interleaving multiple tools for multi-round tool execution to solve complex tasks. Additionally, ToolBench's superior planning and reasoning capabilities are highlighted, as Gorilla does not support multi-step reasoning and fails to execute APIs for obtaining real responses, which is crucial for subsequent model planning. This comparison underscores ToolBench's significance as an improvement over Gorilla.

\medskip Furthermore, the paper indicates the DFSDT's significant outperformance of ReAct \cite{yao2023react} in metrics like win rate and pass rate, showcasing its effectiveness in enhancing decision-making capabilities. The dataset provided by ToolBench is distinctive, focusing on multi-tool scenarios, unlike other datasets like APIBank \cite{li2023apibank} and ToolAlpaca \cite{tang2023toolalpaca}, reinforcing its importance in assessing the generalisation performance of tool-augmented LLMs. 

\medskip Traditional methods attempt to address task decomposition by breaking a task into a chain of sub-tasks \cite{wei2023chainofthought}. However, these approaches rely on the assumption that each sub-task has at most one preceding task, a limitation for real-world applications, especially in multi-modal scenarios requiring multiple inputs. The \textbf{\textit{Tree of Thoughts}} \cite{yao2023treeofthoughts} (ToT) paradigm leverages LLMs for task planning, with edges dynamically formed by LLMs at runtime. On the other hand, the \textbf{\textit{Thoughts-on-Graph}} \cite{2023controlllm} (ToG) paradigm explores solutions on a pre-built graph that captures tool dependencies, mitigating the hallucination problem in tool invocation. The graph's nodes represent tools interconnected based on their dependencies and relationships. ToG overcomes LLMs' token limitations during task planning by searching the optimal solution path on the tool graph instead of relying on LLMs to generate solutions, making it adaptable to changing toolboxes without retraining LLMs. Traversing the graph involves employing a depth-first search (DFS) algorithm, where the tool selection function F samples tool nodes on the graph. The algorithm concludes upon reaching the expected output node or exceeding a set maximum length limit, returning all discovered solutions as a list of tool sequences.

Additionally, the \textbf{ControlLLM}\cite{2023controlllm} paper  introduces several novel modules to manage distinct stages of the answering process: (a) \textit{Solution Expert}: Chooses the optimal solution from the available possibilities, (b) \textit{Solution Description Formatting}: Converts ToG output to string input, and (c) \textit{Solution Evaluation}: Utilises LLMs to evaluate solutions, employing prompt engineering to compare them with subtask descriptions and selecting the solution with the highest score.

\medskip The \textit{\textbf{R}easoning-vi\textbf{a}-\textbf{P}lanning \textbf{(RAP)}} paper \cite{hao2023rap} tackles LLMs' struggle with complex reasoning tasks. They propose an internal \emph{world model} to predict the \emph{world state} and simulate long-term outcomes of actions. The LLM incrementally builds a reasoning tree under given reward metrics. A modified version of Monte Carlo Tree Search is then employed to obtain a high-reward reasoning path. We extensively incorporate their ideas in RE-GAINS, shown in Section \ref{Subsection: Heavier Pipeline}.

\subsection{Benchmarks}
Benchmark datasets are critical for the standardised evaluation of models, offering an impartial platform to measure and compare performance across various approaches within a certain field. \textit{API Bank} \cite{li2023apibank}, a benchmark specifically designed for tool-augmented LLMs, utilises both accuracy and ROUGE scores for its evaluation metrics. While \textit{API Bank} is effective for scenarios where a single tool corresponds to each query, and thus accuracy is a suitable metric, this is not entirely applicable to our context where multiple tools may be required.

Another significant benchmark, \textit{ToolBench} \cite{xu2023toolbench}, encompasses an array of software tools for practical tasks, utilising a success rate as its evaluation criterion. The metric is determined by executing the actual APIs and verifying the results. However, this evaluation method is unfeasible for our research since direct access to the APIs for such execution is beyond our reach.

The \textit{ToolQA} benchmark \cite{zhuang2023toolqa} presents an innovative, automated approach to dataset curation, which leverages specialised tools for engaging with external knowledge in question-answering (QA) tasks. Importantly, \textit{ToolQA} is adept at testing a system's intrinsic logical reasoning capabilities when using such tools.

In \textit{APIBench}\cite{patil2023gorilla}, a novel evaluation method involving Abstract Syntax Tree (AST) matching has been adopted. This technique involves transforming generated code into an AST and analysing the APIs invoked and their arguments. In contrast, our work has identified a more straightforward evaluation solution that better aligns with our methodology, details of which will be provided later in the report.

Additionally, \textit{ToolAlpaca} \cite{tang2023toolalpaca} evaluates using GPT-4, comparing the predicted solution to the actual solution to assign a score. This approach offers another dimension through which we can assess the accuracy and effectiveness of tool-assisted language models.

\subsection{Limited Tools Usage} Various sources \cite{xu2023toolbench, schick2023toolformer, chen2023chatcot, zhuang2023toolqa, lu2023gear, patil2023gorilla} highlight a recurring limitation in the implementation of these techniques—they are confined to a fixed and restricted set of tools. The constraint of limited tool usage raises concerns about the adaptability and scalability of existing techniques in diverse tool-enriched environments.


\subsection{Embeddings for Tool Retrieval} A consistent pattern is observed across multiple references \cite{chen2023chatcot, paranjape2023art, li2023apibank, xu2023toolbench}, where a sentence embedding model is employed to assess the semantic similarity between a tool's description and the query. The tool that aligns most closely with the query is selected for the task. While this method proves effective for simple tasks requiring a single tool, a complex query requiring several tools, presents challenges for tool selection using the embedding model. We perform a pilot experiment using the MiniLM-L6-v2 \cite{reimers-2019-sentence-bert} and the all-mpnet-base-v2 \cite{all-mpnet-base-v2} models, known for their specialisation in tasks like clustering or semantic search, yielded sub-optimal results for our requirements.  


\subsection{Agent-Based learning}
\textbf{The Experiential Learning} (\textit{ExpeL}) framework \cite{zhao2023expel} pioneers methodologies for learning from experiences without necessitating parametric updates. This framework enables the ExpeL agent to engage with training tasks autonomously, extracting and applying knowledge expressed in natural language. The agent attempts problems, reviews correct solutions afterwards, and records insights for future inference, proving its learning efficiency and performance improvement as it accumulates experiences. This process is augmented by reinforcement learning techniques, which optimise the generation and application of these insights.

Another salient open-source framework is \textit{AutoGPT} \cite{autogpt}, focusing on generating logical agents and managing interactions with them. Its widespread popularity and impressive endorsement by approximately 152k users underscores the significance of agent-based learning, a concept we extensively adopt later this report.

Additionally, the paper \textit{"Cognitive Architectures for Language Agents"} \cite{sumers2023cognitive} explores the constituent elements necessary for an ideal cognitive LLM Agent. It delineates components such as various types of memory, action sets, and learning mechanisms, thereby contributing a pivotal blueprint for structuring cognitive capabilities in language models.

\subsection{Discussion on Latency and Cost}

Our empirical observations revealed that fine-tuned GPT-3.5 Turbo exhibited considerably lower latency, achieving speeds up to 3 times faster. It can be primarily attributed to its reduced input/output tokens, at the expense of a threefold increase in inference cost than GPT 3.5. Despite this cost escalation, the performance enhancement is quite notable. We evaluate open source models such as OpenChat \cite{openchat} and Zephyr-7B \cite{tunstall2023zephyr} to further mitigate costs on platforms like Replicate. Additionally, to reduce latency, we explore advanced tactics such as \textit{Paged Attention }\cite{pagedattention} with the \textit{vLLM} \cite{vllm} library. It promises significant speed enhancements, potentially up to 24 times.


\subsection{Discussion on Hallucination}

In the realm of complex question answering, hallucination remains a particularly prevalent challenge. The phenomenon is extensively documented in the paper "ToolQA" by Zhuang et al. \cite{zhuang2023toolqa}, which illustrates instances of hallucinations by the ReAct model, based on GPT-3.5 \cite{yao2023react}, when tasked with QA exercises. Similarly, the creators of the Gorilla framework \cite{patil2023gorilla} indicate that GPT-4, especially when implemented via Hugging Face, suffers from severe hallucination issues. However, they propose that Gorilla significantly mitigates this problem.

Factors to be vigilant about include:

\begin{itemize}
    \item Ambiguous queries that lack clarity in task specification or involve an incorrect sequence of operations.
    \item Queries that introduce many arguments may lead to difficulties, as LLMs have exhibited limitations in handling extensive argument sets effectively.
    \item Consideration should also be given to queries necessitating mathematical logic or reasoning, akin to the higher-level cognitive tasks highlighted in the bonus section of related literature.
\end{itemize}

This leads us to LLM Enforcers which solve a major issue with Language Models, especially in the context of Multi-Agent Systems (Section \ref{Subsection: MultiAgent}). When requiring a precise output format, LLMs do not always perform as instructed. Prompt engineering techniques are not always sufficient. Output enforcers filter the model's generated tokens at every time step. Possible Solutions found to control outputs. We leverage  \textit{\textbf{ToolDec}} \cite{anonymous2023tooldec} and \textit{\textbf{LM Format Enforcer}} \cite{lm-format-enforcer} heavily in EnChAnT (Section \ref{subsec: llm hallucination})

\subsection{Ideas on Data}
It is generally agreed that creating tool-based language learning models starts with producing a high-quality dataset, which is then used to fine-tune an LLM like LLaMA \cite{touvron2023llama} or Vicuña \cite{zheng2023judging-vicuna}. This approach is supported by multiple papers \cite{tang2023toolalpaca, parisi2022talm, patil2023gorilla, xu2023toolbench}. Training with simulated data has been shown to effectively prepare models for specialized real-world scenarios involving the use of tools \cite{tang2023toolalpaca}. In the case of models like GPT 3.5, optimal performance is typically achieved by providing a range of 50-100 examples \cite{ft_gpt_article}. Our efforts are directed toward the automated creation of refined datasets, prioritizing the quality of data over its quantity. Further details of our methodology can be found in Section \ref{Subsection: Data Generation}.

\subsection{Data Generation}
\label{Subsection: Data Generation}


We aim to generate high-quality data and train our agents to make inferences.
This involves the simulation of a software company, including the staff and users. Our contribution generated about 200 distinct tools across 1800 fields ranging from `Data Analytics Tools' to `Text Editing or Word Processing Software'. Our optimized code structure resulted in a cost of only about \$3.5 using the GPT-3.5 Turbo API to create all the tools. We follow three methods for the generation of tools:

\paragraph{Multi-Agent Framework}
\label{Subsection: MultiAgent}

The use of multi-agent systems has shown promise in automatically generating large volumes of training data \cite{xu2023expertprompting, talebirad2023multiagent, autogpt, wu2023autogen, weng2023prompt-Autonomous-Agents}. API Bank and ToolAlpaca have highlighted the importance of agent self-perception in improving task performance \cite{li2023apibank, tang2023toolalpaca}. In our implementation, we utilized a multi-agent framework for autonomous data generation. This involved a simulated group dialogue among agents with different areas of expertise, akin to a corporate hierarchy. A critique agent refined the outputs to maintain high-quality results while significantly reducing the cost of annotation by 98\% compared to human annotation \cite{li2023apibank}. However, we encountered difficulties in standardizing the format of the model outputs, which hindered effective processing and agent interaction.

\paragraph{Conversation Framework}
\label{Subsection: MultiAgent}

Using the RAP and ControlLLM prompt techniques that are discussed later in the paper on GPT-4, we have generated an effective query generator. It converses with the solution generator and, with human feedback, produces high-quality data. We make our "golden" evaluation dataset with this.

\paragraph{Experiment A: Flow Based Generation}
Our first attempt at data generation was specified hand-crafted domains and respective end-users. Domains may be Product management, Project management, Software Architecture etc. We used these domains and a custom prompt to generate \textit{flows}. We define flows as probable scenarios with domain and end-user as guiding inputs. A flow is a JSON object with action items action descriptions, and dependencies between actions connecting based on the order/timeline of actions concerning their input-output correspondence. 

Further, we curate these flows using a custom tree-based approach to account for similar flows or redundant action items by backtracking and using the DFS (Depth First Search) algorithm. We utilized these flows further to generate well-documented step-by-step query descriptions. These serve as our guiding tools for generating solutions. Finally, we feed these flows along with query descriptions to GPT Models to generate the ideal solutions, which are nearly 100\% accurate in utilizing the ControlLLM prompt. Also, we generate human-like concise queries from step-by-step descriptions to compile the final training data with queries and their JSON solutions. Some limitations we faced in this approach were hallucination in entity information and argument values information. Also, the generated examples lacked diversity in query-solution pairs, with instances of repetition and quite unrealistic queries.

\paragraph{Experiment B: Persona Based Generation}
Here, we explain how our final data generation pipeline tackles the limitations we faced in our previous approach. We start by hand-crafting multiple topics/domains which require API frameworks such as "Sales and Marketing", "Cybersecurity and Data Protection", "Inventory Management", etc. We define 18 such independent domains. Next, we prompt the GPT-3.5 turbo model to describe 5 possible personas working with the respective software for all the domains mentioned above. We extensively experimented with various models like Zephyr-7B, OpenChat-13B, Llama-13B and GPT-4-1106-preview and found GPT-3.5 turbo performed the best, with the right balance between performance and cost. We enable the model to generate additional relevant information about various personas, such as their age, profession, education, hobbies, etc., to better generalize on Out-of-Distribution data with increased diversity. 

Next, we prompt GPT to generate all the relevant entities required in the respective domains and their properties. We then generate possible states based on the specific domain and associated entities, representing potential real-life scenarios in the software pipeline. Furthermore, we generate a 5-level task, distributed into five micro-actions based on the given state, each accompanied by a detailed, elaborate description.


\subsection{Graph-of-Thoughts}
We explored the Graph of Thoughts (GoT) framework for modelling thoughts \cite{besta2023got}. In GoT, thoughts are represented as vertices in a graph, and dependencies between thoughts are represented as edges. This allows for aggregating related thoughts by constructing vertices with multiple incoming edges. It is said GoT can extend the capabilities of existing frameworks, such as the Chain of Thoughts (CoT) and Tree of Thoughts (ToT), to accommodate more complex thought patterns. 



\subsection{Stepback Prompting}
The RAP paper \cite{hao2023rap} highlights step-back prompting as a key element in achieving SOTA outcomes in mathematical reasoning tasks. 



\section{Experimentation}

\subsection{New Prompting Techniques}
In one of our proposed solutions, EnChAnT (discussed in Section \ref{Subsection: Efficient Pipeline}), it is necessary to prompt an LLM such as OpenChat. We incorporate insights from prompting strategies we examined from earlier experiments. Specifically, to decompose a user query, our designed prompt instructs the model to formulate a thought considering the subsequent step. The few-shot prompting method is the most effective, balancing efficiency and performance. Additional details can be found in Section \ref{Subsection: Efficient Pipeline}.

\subsection{Graph-based Type Checking}
In our work, we reference the output of the ith tool using the tag \texttt{"\$\$PREV[i]"}. While experimenting with different LLMs and prompts, we noticed that the LLM was hallucinating the substitution of argument values. For example, in place of \texttt{"\$\$PREV[i]"} the LLM may hallucinate \texttt{["\$\$PREV[i]"]} and vice-versa. 
We developed a Graph-based algorithm for type-checking these errors. The algorithm works as follows - \\
\begin{itemize}
    \item \textbf{Initialisation:} Using the Tool database, we build a directed graph with nodes as tools and directed edge with weight $1$ from tool $1$ to tool $2$, if the output of tool $1$ can be directly fed into the input of tool $2$. Similarly, a directed edge with weight $2$ is drawn if the output of tool $1$ is fed into the input of tool $2$ within a list.
    \item  \textbf{Type-Checking:} To check whether the \texttt{\$\$PREV[m]} is compatible tool $N$'s arguments, the algorithm checks if there is a edge between tool $N$ and tool m with the required arguments types. It also updates the arguments if the \texttt{\$\$PREV[m]} requires to be passed as an array
\end{itemize}

\subsection{Reflexion}



Reflexion represents a recent advancement in LLMs, mitigating hallucination in generative models. The architecture incorporates a feedback loop, resembling an LLM-in-the-loop rather than a human-in-the-loop approach. It converts binary/scalar feedback from the environment to text, acting as a `semantic' gradient descent. Reflexion introduces a system with three models: the Actor, Evaluator, and Self-Reflection Model. The Actor generates output from the initial query, the Evaluator provides feedback akin to an environment responding to actions, and the Self-Reflection Model transforms this feedback into questions for the agent. The iterative process, resembling human learning, involves the agent generating responses based on self-evaluator questions, enhancing performance by learning from prior mistakes. In the ExpeL paper \cite{zhao2023expel}, the Self-Reflection agent generates cues during operation, which results in better performance than manually given instructions. An example cue is:
\noindent\begin{minipage}{.45\textwidth}
\small
\centering
\begin{lstlisting}[caption=Example of cue generated by Self-Reflection Agent ,frame=tlrb]{Name}
The actor included unnecessary tools in
the solution, such as  '  ' and  '  '. 
These tools are not required to 
determine the priority of the 
objects related to aorganizationon. 
The actor did not use the '   ' tool, 
which is not necessary for this query, 
but it should have been used to ensure 
the current user is properly identified. 
Revised solution:
\end{lstlisting}
\label{listing: Reflexion Cue}
\end{minipage}\hfill



\subsection{Output Enforcer for GPT!}

\begin{figure*}[ht]
    \centering
    \includegraphics[width=0.8\linewidth]{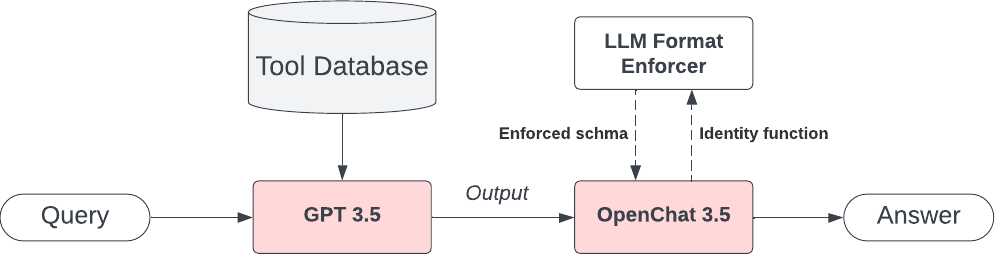}
    \caption{A novel pipeline to use LLM enforcers on closed-sourced models like GPT 3.5}
    \label{fig:llm enforcer}
\end{figure*}

Observing the success of EnChAnT (Section \ref{Subsection: Efficient Pipeline}), we note how using an LLM Enforcer reaped great results. A simple OpenChat model and the LLM enforcer ensured unparalleled results. We propose a novel idea to extrapolate this idea to GPT 3.5. The LLM enforcer is only available for use on open-source models since the enforcer manipulates the token prediction probability of the model's final layer. We extend this to closed-source models. Figure~\ref{fig:llm enforcer} shows the same in pictorial form. To ensure we reduce hallucinations on models such as GPT 3.5, we propose the following -
\begin{itemize}
    \item Get output from a closed-source model like GPT 3.5.
    \item Pass this output as an input to an open-source model like OpenChat. 
    \item Prompt the OpenChat model to exactly replicate the inputs as its inputs - similar to an identity function.
    \item Use LLM Enforcers on the OpenChat model. Since the OpenChat model is asked to give out the same output as its input from GPT, we are effectively enforcing the outputs of a closed-source GPT 3.5 model.
\end{itemize}


\section{A comparison of LLMs}
\label{Section: Comparision of LLMS}

In our evaluation, we noted GPT-4-turbo's exceptional performance in accurately answering every question posed by our problem statement. GPT-4-turbo's proficiency made us consider it a reliable source for establishing ground truth for newer data samples. We also leverage it to generate an extensive tool dataset. Despite the allure of GPT-4-turbo's capabilities, we must contend with its high token cost. Hence, we pivot towards more cost-effective solutions such as finetuning GPT-3.5-turbo or opting for accessible open-source alternatives like LLama, Zephyr, and OpenChat hosted on remote servers. This ensures our project progresses with efficient use of computing resources at a lesser cost.

\begin{table*}[b]
    \centering
    \resizebox{2\columnwidth}{!}{
        \begin{tabular}{lcccc}
        \toprule
        Model &Type & API cost (input/output) & latency(input/output)&CAS \\
        \midrule
        OpenChat \cite{openchat} & open-source &\cellcolor{yellow!35} 0.0006 /0.0095& 0.8373/0.8105 & - \\
        Zephyr-7B \cite{tunstall2023zephyr} & open-source &\cellcolor{yellow!35} 0.0009/0.009& 0.7705/1.2603&0.71 \\
        GPT-3.5 Turbo  & proprietary & 0.001/ 0.002& 0.4963/12.5231 &0.75 \\
        GPT-4 \cite{openai2023gpt4} & proprietary & 0.01/0.03 &\cellcolor{yellow!35}0.01/45.3934& 0.76 \\
        GPT-3.5 Turbo fine-tuned & proprietary & 0.003/0.006 & \cellcolor{yellow!35}0.0497/9.3503&-  \\        
        \bottomrule
        \end{tabular}
    }
    \caption{Comparison of LLM Cost is in \$/1K tokens.The empty values are not available at the moment. Latency is in sec/1000.Context Adherence Score-CAS}
    \label{table: compare llms}
\end{table*}



We checked other models such as Llemma \cite{azerbayev2023llemma} because of its high reasoning ability and grasp on mathematical ability. Even though it has been established specifically in code writing skills, we look at it through the lens of result evaluation. We find that GPT-4 still emerges as a more powerful model than Llemma! 
\footnote{H2O allows users to seamlessly compare outputs from multiple LLMs \url{https://gpt.h2o.ai/}}

\section{Evaluation Metrics}
\label{subsec:evaluation-metrics}
We adopt a multi-faceted evaluation framework inspired by ControlLLM \cite{2023controlllm}, which considers tool selection, argument assignment, and overall solution adequacy. Below, we outline the same

\subsection{Tool Selection Metrics}
\begin{itemize}
    \item \textit{Irrelevant Tool Inclusion Rate (IR)}:
    Ratio of irrelevant tools to predicted tools.
    \item \textit{Necessary Tool Inclusion Rate (NR)}: Ratio of necessary tools to predicted tools.
    \item \textit{Missing Tool Rate (MR)}: Ratio of missing tools to necessary tools. 
\end{itemize}

\subsection{Argument Assignment Metrics}
\begin{itemize}
    \item \textit{Resource Hallucination Rate (HR)}: Identifies the prevalence of nonexistent resources in the tool arguments provided by the method. Lower HR values suggest a reduced tendency towards creating hallucinated arguments.
\end{itemize}

\subsection{Solution Evaluation Metrics}
\begin{itemize}
    \item \textit{BLEU Score}: Assess similarity between generated outputs and actual target.
    \item \textit{Rouge-L F1 Score}: measures the longest common subsequence (LCS) between the system-generated solution and the reference solution. The F1 score computes the harmonic mean of precision and recall, where precision is the ratio of the length of the LCS to the length of the system solution, and recall is the ratio of the length of the LCS to the length of the target solution. 
\end{itemize}

\subsection{Efficiency Evaluation Metrics}

\begin{itemize}
    \item \textit{API Call and Token Count}: We tally the quantity of API calls and token generation to approximate the cost implications.
    \item \textit{Correct Path Rate:} Model can generate a solution path which may contain the most optimum solution as a subsequence , such solutions these paths are used to calculate the Correct Path Rate of the model.
    
\end{itemize}

The metrics discussed find precedence in several key works \cite{qin2023toolllm, song2023restgpt, 2023controlllm}, and have informed our selection of Irrelevant Tool Inclusion Rate (IR), Missing Tool Rate (MR), and Resource Hallucination Rate (HR).





\section{Benchmarking}

\subsection{ControlLLM}
ControlLLM \cite{2023controlllm} proposes a Thoughts-on-Graph based method for tool selection, aiming to enhance scalability and prevent hallucination. In our implementation, we adopt the task decomposition aspect of the model utilizing their prompting technique to break down user queries into sub-tasks. This implementation with GPT-4 achieves 100\% accuracy, as detailed in the provided prompt found in Appendix \ref{appendix: ControlLLM Prompt}. However, a limitation is observed due to the high cost associated with this approach, the large number of tokens required and the expense of GPT-4.

\subsection{RAP}
Like ControlLLM, we observe that the RAP paper, when used in conjunction with GPT-4, gives a near-perfect performance. Hence, we leverage RAP for benchmarking our model's outputs. Besides this, we also found that using a refined version of RAP instructions helped GPT-3.5 turbo generate very good results without any training. This forms our final pipeline.


\subsection{Retrievers}

In our methodology, we harness the capabilities of two distinct retrievers, first is the OpenAI’s text embedding 'openai/text-embedding-ada-002' and the second is 'Toolbench IR\_bert\_based\_uncased' (ToolBench's API Retriever)\cite{xu2023toolbench}, which serve as crucial components within our tool recommendation system. The first retriever, 'openai/text-embedding-ada-002,' retriever is used in our first proposed pipeline, RE-GAINS. On the other hand, the second retriever, based on BERT and employed in the EnChAnT pipeline, is a sentence transformer model. It maps sentences and paragraphs to a 768 dimensional dense vector space and is useful for tasks like clustering or semantic search. We evaluate the retrievers for their efficiency in Appendix Table  \ref{table: rap-eval}


\subsection{Finetuning GPT 3.5}
Our initial observations indicated that the vanilla GPT 3.5 model often produced 'hallucinated' results, occasionally disregarding the predefined JSON schema, leading to invalid JSON outputs. Such discrepancies were not rare, including cases where the model would omit necessary tools or select inappropriate ones.

To enhance performance and reduce such errors, we fine-tuned GPT 3.5 with 35 gold-standard samples, resulting in a significantly faster model than its base counterpart. This fine-tuned model was specifically trained to generate outputs capitalising on Python's ability to produce more compact code than verbose JSON structures. 

Besides improving response speed and output clarity, this conversion to Python also targets cost efficiency. As outlined in Section \ref{Section: Comparision of LLMS}, output tokens incur greater expense than input tokens. optimising output length is crucial. By adopting the GPT-3.5-turbo pricing model, we cut response time by about 15\% 
of 0.0457 after training for 4 epochs on our custom Gold-standard dataset. Upon testing the finetuned model after adding new tools, we found that it did not lose its generalisation ability. We show this in Appendix \ref{Measuring efficiency of Typescript}.

When contrasting our fine-tuned GPT 3.5 with open-source models, a stark difference in performance emerges. The latter tend to lag in terms of time complexity, taking around 6 seconds to generate output—substantially slower than what was achieved by our fine-tuned version of GPT-3.5. Comparative experiments with OpenChat \cite{openchat}, while yielding slightly slower processing times, still demonstrated the utility of our TypeScript-JavaScript approach. This innovative methodology conserves token usage and accelerates response times, making it a game-changing adaptation for our project's interaction with LLMs. 


\subsection{On Manipulation of tools}
The versatility of our model shines in handling output manipulation. The Python-based outputs are useful, particularly due to Python's support for operator overloading. By overloading operators within the output classes, we can seamlessly carry out various operations on the results. This feature enables us to embed complex logic within these operations effortlessly. 

Given the straightforward implementation of logical operations combined with Python's capacity to manage primitive data types effectively, we decided to exclusively utilise Python for managing our outputs. This decision ensures that every operation, from arithmetic such as addition and subtraction to comparison operators like greater than and less than, are easily implementable.

To accommodate the manipulation of outputs, we incorporate functionality to perform arithmetic operations—addition, subtraction, division, and multiplication—and more advanced operations, including floor division, exponentiation, and modulus. Moreover, we handle comparison operations, ensuring our system accurately assesses greater than, less than, equality, and inequality. These operations are coded directly into the system, streamlining the manipulation process and solidifying the effectiveness of using Python for our project's output management needs.







\section{Proposed Solution: Retrieval Enhanced Generation via Actions INsights and States (RE-GAINS)}
\label{Section: Arya Pipeline}

\begin{figure*}[ht]
    \centering
    \includegraphics[width=0.9\linewidth]{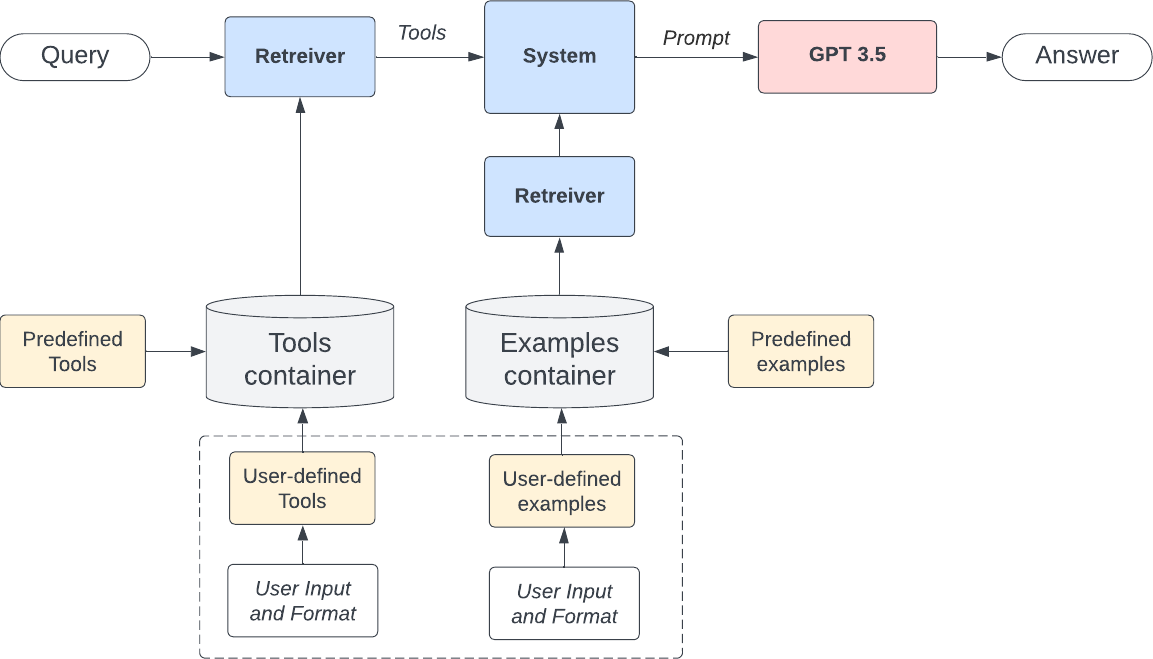}
    \caption{Pipeline for the main proposed solution. The dotted lines represent the optional component of the process.}
    \label{fig:arya pipeline}
\end{figure*}

Our proposed solution utilizes an effective tool and example retrieval system coupled with a novel prompting technique to effectively solve the query using just a single LLM API call. The process involves tool retrieval, task decomposition and step-by-step reasoning. This is the best solution, taking deployment cost and inference times into consideration.
\subsection{Salient Features of our proposal}
Retriever chooses the relevant tool to solve the problem. Then, examples retriever chooses relevant examples which use the tools. This speeds up the reasoning process and reduces input tokens. A novel prompting technique utilizes just a single API call with relatively small number of input tokens. This minimizes cost drastically and makes reduces latency significantly. Examples from our "golden" dataset are augmented to increase accuracy. Users can freely add new tools or add relevant examples. This approach is extremely robust and generalizable and works exceedingly well on a one-shot setup.

\subsection{Retrieval}
Both tools and examples superset is embedded in Ada embeddings. 

\subsection{Novel Prompting Techniques}
From the paper on Reasoning via Planning (RAP) \cite{hao2023rap}, we learnt a few critical things. Firstly, we introduce the concepts of "state" and "action". This helps in task decomposition step-by-step by asking sub-questions about which tool(action) to use to get the next "state". Many "actions" can be taken at any state. The LLM evaluates each one of them to decide the next action, which ensures it covers a larger part of the reasoning space. Further, the LLM maintains a "world model" that maintains context concerning the task itself and how any new "action" may affect the overall picture. This greatly eliminates hallucinations.

From the Expel paper \cite{zhao2023expel}, we noticed that the best way to decide what the next state should be is a series of "insights". These provide the LLM with a set of guidelines that allows it to choose the correct action with minimal hallucination. We carefully curate these insights by analyzing how the sub-questions were developed and answered by RAP-prompted GPT-4.

Our prompt utilizes the best ideas of both papers. The effective task decomposition of RAP combined with good quality "insights" to make decisions at the task level makes this model highly accurate. The prompt can convey such a large amount of information and maintains a relatively low number of input tokens(~2900 tokens for 17 tools and two large examples). Further, it is highly generalizable and works well on new tools without adding relevant examples.

\subsection{Justifying use of OpenAI model and embeddings}
Firstly, GPT-3.5-turbo performed fairly poorly on most tasks on its own and in implementations of many recent papers. The model is cheaper to deploy cost-wise than the much more powerful GPT-4. On the other hand, deploying "open-source" models on platforms like Replicate is costlier in this application, as it bills time-wise. Hence, implementing our pipeline on GPT-3.5-turbo makes perfect sense as it is very cheap and has fast inference times. On average, inference of large complex queries from our "golden" dataset takes less than Rs 0.4.

The Ada embeddings are very cheap and effective and are better than hosting open-source retrievers on a cost basis. Per query, the costs are almost negligible.



\section{Proposed Solution: Enforced Creation of Actions and Thoughts (EnChAnT)}
\label{Subsection: Efficient Pipeline}



\begin{figure*}[ht]
    \centering
    \includegraphics[width=0.9\linewidth]{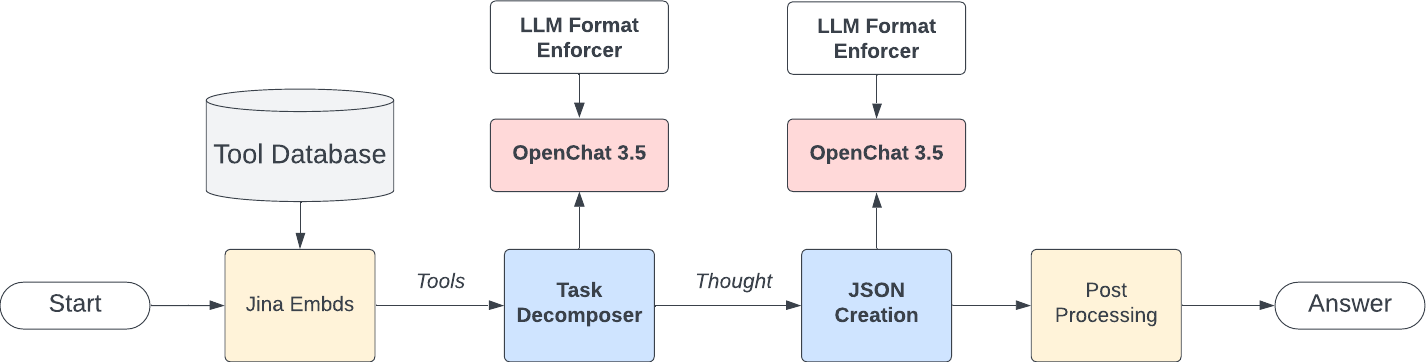}
    \caption{Pipeline of Proposed Architecture}
    \label{fig:mahendra-pipeline}
\end{figure*}

This model leverages the usage of ToolBench's API Retriever to optimize tool retrieval \cite{xu2023toolbench}. The system is structured around a tailored three-stage process comprising Tool Retrieval, Decomposition, and Recomposition, designed with a focus on maximizing efficiency in terms of both time and cost metrics.

\subsection{Salient Features of our Proposal}
Noteworthy features of our pipeline include full open-source implementation the complete elimination of tool and argument hallucinations through the use of the Language Model Format Enforcer (LM Format Enforcer). Integrating the open-source OpenChat into our chat model is a pivotal decision; its cost-efficient deployment in Replicate adds operational efficiency, optimising resource management. Even with a 4-bit quantised model, we accomplish high-quality task decomposition. The entire model can be run with less than 6 GB GPU. Due to the enforcement of tool and argument names, we get a negligible hallucination rate, showcasing the benefits of controlled generation.

\subsection{Discussion on LLM Hallucination}
\label{subsec: llm hallucination}
Implementing the Language Model Format Enforcer (LMFE) is crucial in mitigating tool name hallucination. LMFE  systematically reduces the probabilities assigned to disallowed tokens in the output space to zero. Extending beyond simple JSON format enforcement, complex JSON schemas can be enforced. This compels the system to prioritize and utilize only permissible tokens, hence eliminating hallucinations.


\subsection{The Pipeline:}
\paragraph{Step 1: Tool Retrieval}
We use ToolBench's API retriever (ToolBench\_IR\_bert\_based\_uncased) to convert the query and tools from the text format to Embeddings. Then, we take the cosine similarity between the tool embeddings and the query to get an array with similarity scores for each tool concerning the query. From this, we pick the top-k (10, in our case) tools. 

\paragraph{Step 2: Task Decomposition}
We few-shot prompt an open-chat model \cite{wang2023openchat} to decompose the given query into multiple sub-tasks, each associated with one tool for resolution. A custom structure for these sub-tasks has been developed, and its adherence is enforced by the LMFE. Additionally, the LMFE eliminates tool name hallucinations. Prompt is given in Appendix \ref{appendix: TASK DEC}
\paragraph{Step 3: Task Re-composition}
We again use a one-shot prompt with open-chat \cite{wang2023openchat} model to recompose the given sub-tasks into one cohesive JSON file. We use LMFE to enforce the JSON schema. The LMFE also eliminates the tool name hallucinations and argument name hallucination. We do some post-processing on the output from the open-chat, to remove common mistakes.

\subsection{In essence}
RE-GAINS can effectively work for the first time on a new set of tools with minimal hallucination. This is because of the highly generalizable and robust prompt coupled with the retrievers that always get relevant examples and tools for good context. Further, the tool generates largely accurate solutions for queries that can only be partly solved correctly.

\begin{table*}[ht]
\centering
    \resizebox{1.3\columnwidth}{!}{
    \begin{tabular}{lccccccc}
    \hline
        \textbf{Implementation} & \textbf{Model Name} & \textbf{Latency (s)}& \textbf{Cost (\$)} \\ \hline
        RAP prompt & gpt-4-1106-preview & 17.03 & 0.080 \\ 
        \textbf{Re-GAINS} & gpt-3.5-turbo & \cellcolor{yellow!35}8.62 & \cellcolor{yellow!35}0.009  \\
        \textbf{EnChAnT} & openchat-3.5 & 35.68 & \cellcolor{yellow!35}0.008  \\ 
        \hline 
        \end{tabular}}
        \caption {For a similar number of tokens (\~2600  input/260 output)}
    \label{table: rap-eval}
\end{table*}

\section{Evaluation}
We evaluate our experimented models on the metrics defined in Section \ref{subsec:evaluation-metrics}. We show benchmarking results of expensive-yet-strong models like GPT4 with Control LLM and GPT4 with RAP in Table \ref{table: benchmarking eval}. We observe that RAP prompt with 'gpt-4-1106-preview' model gives the most accurate results.
We then compare the results of our RAP-based GPT3.5 model, with other RAP-based models. This is shown in Table \ref{table: rap-eval}, which highlights the efficiency of our proposed solution in Section \ref{Section: Arya Pipeline}. Comparing the ControlLLM prompt with RE-GAINS (RAP), we see that the RE-GAINS based on the RAP prompt gives a much lower IR score and a much higher NR and BLEU score. 
\begin{table*}
\centering
    \resizebox{\linewidth}{!}{
    \begin{tabular}{lcccccccc}
    \hline
        \textbf{Implementation} & \textbf{Model Name} & \textbf{IR} $\downarrow$ & \textbf{NR} $\uparrow$ & \textbf{HR} $\downarrow$ & \textbf{MR} $\downarrow$ & \textbf{BLEU Score} $\uparrow$ & \textbf{ROUGE-L-F1 Score} $\uparrow$ & \textbf{Invalid JSON}$\downarrow$ \\ \hline
        ControlLLM Prompt & gpt-4-1106-preview & 0.169 & 0.831 & 0.401 & 0.171 & 0.763 & 0.638 & 0.014 \\
        ControlLLM Prompt & gpt-3.5-turbo-1106 & 0.297 & 0.703 & 0.47 & 0.392 & 0.663 & 0.588 & 0.232 \\
        RAP Prompt & gpt-4-1106-preview & 0.028 & 0.972 & 0.0 & 0.058 & 0.906 & 0.865 & 0.493 \\
        RAP prompt & openchat-3.5 & 0.438 & 0.563 & 0.875 & 0.875  &  0.278 & 0.125 & - \\ 
        RAP prompt & llama-70b & 0.596 & 0.404 & 0.254 & 0.456 & 0.456 & 0.332 & - \\
        
        
        \rowcolor{black!10}\textbf{RE-GAINS} & gpt-3.5-turbo-1106 & 0.039 & \textbf{0.961} & \textbf{0.003} & \textbf{0.164} & 0.780 & \textbf{0.772} & 0.014 \\
        
        \rowcolor{black!10}\textbf{EnChAnT}&
        openchat-3.5 & 0.305 & 0.695 & \textbf{0.01} & 0.345 &0.629 & 0.552 & - \\
        \hline 
        
        \end{tabular}}
        \caption {Comparing benchmark results from the evaluation of different implementations on our \textbf{"golden dataset"}.}
    \label{table: benchmarking eval}
\end{table*}

\section{Future Work}
\subsection{Reasoning Tree }
\label {Subsection: Heavier Pipeline}

In our approach, we implemented a graph-based method focusing on domain classification. The system includes specific domains, such as an exclusive authentication domain and a termination domain marked by an end\_tool. We utilised the Language Model (LLM) to generate sub-questions for each task, which the retriever then used. The graph structure is established by connecting the initial state to sub-states generated by tools within the authentication domain. Subsequent states are determined by the LLM, considering the context of the current tool and all tools leading to it. The LLM selects the domain for further states. The retriever identifies top-n tools within the chosen domain based on sub-questions generated for that domain. These selected tools become the children nodes in the graph. The process involves assigning probabilities to paths based on the retriever's output, and the desired output data type constrains the LLM. For backtracking, we applied Dijkstra's algorithm, considering the probability of each path. The graph, when generated appropriately, supports a depth-first-search approach where the LLM makes tool choices at each stage, ensuring a systematic exploration of possible solutions. A Monte-Carlo tree method to exploit nodes of the tree will make this method highly optimised.

\subsection{Finetuning LLMs}


Our motivation is to achieve a system that is faster, capable of processing logical tasks, and cost-efficient. This is why we chose to utilize Python over JSON. All tools and their corresponding solutions were converted to Python format to reduce token sizes. We also experimented by converting them to TypeScript to reduce them further. Admittedly, the conversion process from one language to another does involve some degree of complexity. We leverage meta-programming tactics for the same. Being dynamic instead of being based on sub-processes, it works significantly faster and is much more cost-effective. In other words, code is written during run-time. Converting the Python output to our desired JSON format takes only 120 ms on Google Colab. 

We tried fine-tuning many open-source models, including Mistral, Vicuna, OpenChat, and GPT 3.5. We observed that Finetuning open source models led to generalised models. It can be inferred that such large language models only learn to pick up the output template of the given samples. There is no real learning by the model. This can be intuitively understood as a large model with billions of parameters, a size in gigabytes, being asked to fine-tune on only a few bytes of data!

However, in a surprising achievement, fine-tuning GPT 3.5 led to a very strong model. Finetuning GPT-3.5 enabled a significant reduction in response times. Our fine-tuned version improved performance, delivering responses in approximately 818 ms. By adopting the GPT-3.5-turbo pricing model, we cut response time by about 15\% and costs by nearly 37\%. We experimented with a Python toolset to generate JSON outputs to reduce the token cost. However, it was seen that the model performed well on JSON input and outputs compared to using Python for any stage. Our model performs almost perfectly on all available DevRev queries. However, we do not keep the solution as our proposed solution. 
Comparative experiments with OpenChat \cite{openchat}, while yielding slightly slower processing times, still demonstrated the utility of our TypeScript approach. This innovative methodology conserves token usage and accelerates response times. However, we observed that the fine-tuned OpenChat model does not perform comparably to the fine-tuned GPT3.5 model.
\subsection{Domain Classification}
Taking inspiration from the software architecture, we propose automatically generating the objects from the tool descriptions. Then we try to predict the attributes of these entities. In addition, we can automatically organize the given set of tools into static tools and methods grouped by entity. This way we rephrase the given tool manipulation problem into a software architecture-based problem. In our experiments showed that this approach ensures tools like "who\_am\_i" are easily recognized and utilized. This is something that other approaches generally struggle with.

\section{Conclusion}
In conclusion, our project focuses on developing efficient and accurate AI agents for tool-augmented language models. We have proposed two pipelines, RE-GAINS and EnChAnT, aiming to maximize efficiency and minimize cost while ensuring accurate results. 

RE-GAINS incorporates a tool, an example retrieval system, and novel prompting techniques to solve queries using a single LLM API call. This pipeline has been designed to be highly scalable and cost-effective, making it suitable for various domains. 

On the other hand, EnChAnT leverages a state-of-the-art retriever for tool retrieval and implements task decomposition and re-composition techniques to generate accurate solutions. This pipeline also prioritizes efficiency and cost-effectiveness.

We have conducted extensive experimentation and evaluation throughout our project to fine-tune our models and ensure their accuracy. We have also explored various literature on data generation techniques, prompting strategies, and control LLMs to enhance our understanding of the field.

We plan to refine our models by incorporating additional techniques such as reasoning trees and domain classification. We also aim to explore more advanced prompting methods and continue bench-marking our models against existing solutions.

Overall, our project represents a significant step towards developing efficient AI agents that can effectively handle language-based tasks with the assistance of external tools.

\section*{Acknowledgements}
We would like to acknowledge DevRev for defining the original problem statement \footnote{AI Agent 007: Tooling up for Success, a problem statement in the Inter-IIT Tech Meet 12.0.}. They also provided us with an initial set of tools and a set of query-solution examples relating to real-world use cases involving tool chaining. 




\end{multicols}

\newpage
\begin{multicols}{2}
    \renewcommand*{\bibfont}{\small}
    \printbibliography
\end{multicols}
\clearpage 

\begin{appendices}

In the appendix section, we provide detailed information on the following aspects of our study

\section{Evaluation of Prompting Techniques}
\label{Section: Eval Prompting Tech}
\begin{table*}[h]
\centering
    \resizebox{1\columnwidth}{!}{
    \begin{tabular}{lccccccc}
    \hline
        \textbf{Prompting Method} & \textbf{Model Name} & \textbf{IR} $\downarrow$ & \textbf{NR} $\uparrow$ & \textbf{HR} $\downarrow$ & \textbf{MR} $\downarrow$ & \textbf{BLEU Score} $\uparrow$ & \textbf{ROUGE-L-F1 Score} $\uparrow$ \\ \hline
        Analogical & gpt-3.5-turbo-1106 & 0.131 & \cellcolor{yellow!35}0.869 & 0.251 & 0.220 & 0.699 & 0.620 \\ 
        Analogical & gpt-4-1106-preview & 0.201 & 0.799 & 0.288 & \cellcolor{yellow!35} 0.061 & \cellcolor{yellow!35}0.752 & \cellcolor{yellow!35}0.676 \\ 
        Analogical & openchat\_3.5 & 0.186 & 0.814 & 0.251 & 0.252 & 0.642 & 0.611 \\ 
        Analogical & zephyr-7b & 0.275 & 0.725 & 0.243 & 0.304 & 0.638 & 0.533 \\\hline 
        CoT & gpt-3.5-turbo-1106 & 0.331 & 0.669 & 0.158 & 0.345 & 0.527 & 0.497 \\ 
        CoT & gpt-4-1106-preview & \cellcolor{yellow!35} 0.083 & \cellcolor{yellow!35}0.917 & 0.288 & 0.055 & 0.769 & 0.706 \\
        CoT & openchat\_3.5 & 0.707 & 0.293 & \cellcolor{yellow!35} 0.110 & 0.759 & 0.194 & 0.221 \\ 
        CoT & zephyr-7b & 0.635 & 0.365 & \cellcolor{yellow!35} 0.111 & 0.625 & 0.334 & 0.334 \\ \hline
        React & gpt-3.5-turbo-1106 & 0.169 & 0.831 & 0.238 & 0.163 & 0.708 & 0.660 \\ 
        React & gpt-4-1106-preview & 0.183 & 0.817 & 0.317 & 0.091 & 0.756 & 0.703 \\ 
        React & openchat\_3.5 & 0.481 & 0.519 & 0.209 & 0.509 & 0.515 & 0.537 \\ 
        React & zephyr-7b & 0.426 & 0.574 & 0.292 & 0.329 & 0.658 & 0.527 \\ \hline
        Stepback & gpt-3.5-turbo-1106 & \cellcolor{yellow!35} 0.121 & \cellcolor{yellow!35}0.879 & 0.244 & 0.144 & 0.741 & 0.611 \\ 
        Stepback & gpt-4-1106-preview & 0.174 & 0.826 & 0.268 & 0.050 & 0.775 & 0.682 \\ 
        Stepback & openchat\_3.5 & 0.168 & 0.832 & 0.282 & 0.327 & 0.634 & 0.601 \\ 
        Stepback & zephyr-7b & 0.313 & 0.688 & 0.148 & 0.215 & 0.596 & 0.562 \\ \hline
    \end{tabular}}
        \caption{Comparing results from evaluation of different prompting methods IR,HR and MR have to be lower,while NR,BLEU score and ROUGE score have to be higher\textbf{(JSON-to-JSON approach)}}
    \label{table: prompting eval-jsjs}
\end{table*}

The following results were obtained for JSON to JSON I/O: Among the models with different capabilities, GPT-4 with CoT shows the best performance in Irrelevant tool Rate (IR), while Zephyr-7B with OpenChat+COT demonstrates the worst performance IR/NR score

For hallucination rate, GPT-4 usually performs worst among the models, but there is not a large difference among the different models. Among prompting methods, CoT performs best.

Missing tool rate is significantly lower for GPT-4 than for other models.

In general, among prompting methods, Stepback prompting gives the best results while CoT performs the worst. Among models, GPT-3.5 is marginally better than GPT-4 in many cases, however GPT-4 shows significantly lower MR.

\begin{table*}[h]
\centering
    \resizebox{1\columnwidth}{!}{
    \begin{tabular}{lccccccc}
    \hline
        \textbf{Prompting Method} & \textbf{Model Name} &\textbf{ IR} $\downarrow$ & \textbf{NR} $\uparrow$ & \textbf{HR} $\downarrow$ & \textbf{MR} $\downarrow$ & \textbf{BLEU Score }$\uparrow$ & \textbf{ROUGE-L-F1 Score} $\uparrow$ \\ \hline
        Analogical & gpt-3.5-turbo-0301 & 0.249 & 0.751 & 0.029 & 0.227 & 0.657 & 0.580 \\ 
        Analogical & gpt-3.5-turbo-1106 & 0.150 &  0.850 & 0.006 & 0.239 & 0.696 & 0.601 \\ 
        Analogical & gpt-4-1106-preview & 0.181 & 0.819 & 0.032 & \cellcolor{yellow!35} 0.046 & 0.605 & 0.623 \\
        Analogical & openchat\_3.5-awq & 0.360 & 0.640 & 0.073 & 0.410 & 0.455 & 0.473 \\ 
        Analogical & zephyr-7b-beta & 0.262 & 0.738 & 0.075 & 0.295 & 0.534 & 0.513 \\ \hline
        CoT & gpt-3.5-turbo-0301 & 0.339 & 0.661 & 0.016 & 0.312 & 0.611 & 0.535 \\ 
        CoT & gpt-3.5-turbo-1106 & 0.186 & 0.814 & 0.019 & 0.288 & 0.633 & 0.569 \\ 
        CoT & gpt-4-1106-preview & 0.265 & 0.735 & 0.017 & 0.180 & 0.593 & 0.591 \\ 
        CoT & openchat\_3.5-awq & 0.704 & 0.296 & 0.090 & 0.730 & 0.272 & 0.297 \\ 
        CoT & zephyr-7b-beta & 0.919 & 0.081 & 0.020 & 0.899 & 0.099 & 0.087 \\ \hline
        React & gpt-3.5-turbo-0301 & 0.639 & 0.361 & 0.008 & 0.681 & 0.313 & 0.280 \\ 
        React & gpt-3.5-turbo-1106 & 0.321 & 0.679 & 0.000 & 0.472 & 0.514 & 0.493 \\ 
        React & gpt-4-1106-preview & \cellcolor{yellow!35} 0.111 & \cellcolor{yellow!35} 0.889 & 0.029 & 0.062 & \cellcolor{yellow!35} 0.735 & \cellcolor{yellow!35} 0.701 \\ 
        React & openchat\_3.5-awq & 0.797 & 0.203 & 0.018 & 0.802 & 0.173 & 0.180 \\
        React & zephyr-7b-beta & 0.808 & 0.192 & 0.087 & 0.788 & 0.238 & 0.207 \\ \hline
        Stepback & gpt-3.5-turbo-0301 & 0.187 & 0.813 & 0.021 & 0.251 & 0.698 & 0.576 \\ 
        Stepback & gpt-3.5-turbo-1106 & 0.183 & 0.817 & \cellcolor{yellow!35} 0.000 & 0.304 & 0.630 & 0.582 \\ 
        Stepback & gpt-4-1106-preview & 0.142 & 0.858 & 0.019 & 0.066 & 0.679 & 0.628 \\ 
        Stepback & openchat\_3.5-awq & 0.602 & 0.398 & 0.013 & 0.622 & 0.233 & 0.282 \\ 
        Stepback & zephyr-7b-beta & 0.592 & 0.408 & 0.045 & 0.552 & 0.331 & 0.310 \\ 
    \end{tabular}}
        \caption{Comparing results from evaluation of different prompting methods .IR,HR and MR have to be lower,while NR,BLEU score and ROUGE score have to be higher\textbf{(TypeScript-to-JSON approach)}}
    \label{table: prompting eval-tsjs}
\end{table*}

\begin{table*}[h]
\centering
    \resizebox{1\columnwidth}{!}{
    \begin{tabular}{lccccccc}
    \hline
        \textbf{Model Name} & \textbf{IR} $\downarrow$ & \textbf{NR} $\uparrow$ & \textbf{HR} $\downarrow$ & \textbf{MR} $\downarrow$ & \textbf{BLEU Score} $\uparrow$ & \textbf{ROUGE-L-F1 Score } $\uparrow$ \\ \hline
        gpt-3.5-finetuned & 0.036 & 0.964 & 0.044 & 0.140 & 0.729 & 0.794 \\ \hline
    \end{tabular}}
        \caption{Performance of fine-tuned GPT-3.5 model for different prompting techniques.This model seems to be performing better than other models observed}
    \label{table: prompting eval-tsjs}
\end{table*}

Among the models with different capabilities, GPT-4 with ReAct/Step-back shows the best performance in Irrelevant tool Rate (IR), while Zephyr-7B with OpenChat+COT demonstrates the worst performance IR/NR score.
\\\\
Missing tool rate is highest for Zephyr-7B with stepback and the lowest is for GPT4 with analogical prompting. Amongst the other prompting techniques, CoT resulted in considerably higher MR for GPT-4. Interestingly, GPT-3.5 Turbo equipped with React/Step Back exhibits notably low hallucination rates, while Zephyr-7B struggles with high hallucination rates and GPT-4 with Step-back/ReAct performs unexpectedly poorly. In terms of Metrics, GPT-3.5 Turbo with Analogical/Step-back/ReAct scores higher on BLEU metrics, while the trend remains consistent for ROUGE scores.
\\\\
CoT seems to affect the performance of models adversely, so much so that, GPT-3.5 with other methods perform better than GPT-4 + CoT. Overall, GPT-4 with Step back and ReAct Prompting Technique seems to be the best performer overall, followed by GPT-3.5 Turbo, with the above mentioned methods.
 Zephyr-7B with CoT Prompting seems to be the worst performer overall, followed by Zephyr-7B with Step back and ReAcT Prompting Technique. Overall, ReAcT and Step Back prompting technique seems to increase the overall score of most models with the exception of ReAcT+GPT-3.5-Turbo-0301
Strangely, ReAct prompting seems to adversely affect the performance of GPT-3.5-Turbo-0301.

\section{Latency for GPT models}
\begin{table*}[h]
    \centering
    \small
    \resizebox{0.6\columnwidth}{!}{
        \begin{tabular}{lc}
        \toprule
        Model Name & Avg Inference Time/50 tokens \\ 
        & (in seconds) \\ \midrule
        gpt-4-1106-preview & 4.851 \\
        gpt-4-0613 & 7.717 \\
        gpt-4-0314 & 7.949 \\
        gpt-4 & 4.911 \\
        gpt-3.5-turbo-16k-0613 & 7.083 \\
        gpt-3.5-turbo-16k & 7.058 \\
        gpt-3.5-turbo-1106 & 3.878 \\
        gpt-3.5-turbo-0613 & 6.00 \\
        gpt-3.5-turbo-0301 & 5.582 \\ \bottomrule
        \bottomrule
        \end{tabular}
    }
    \caption{Output Latency Measurements for OpenAI models from our Experiments. It can be seen that the models with the suffix \texttt{1106} are much quicker than their other counterparts}
    \label{table: latency measurements}
\end{table*}

\section{Evaluating Retrievers}
\begin{table}[hbt!]
\centering
    \resizebox{0.6\columnwidth}{!}{
    \begin{tabular}{lcc}
    \hline
         & \textbf{OpenAI} & \textbf{ToolBench Retriever}\\ \hline
        Top 5 & 0.7625 & 0.7325  \\ 
        Top 7 & 0.8562 & 0.8367 \\ 
        Top 9 & 0.9479 & 0.9362 \\
        \hline 
        \end{tabular}
        }
        \caption{Bench-marking the two major Dense retrievers we use - OpenAI (\texttt{openai/text-embedding-ada-002}) and Jina Retrievers \texttt{ToolBench Retriever}. The Top 'N' score indicates the average percentage of tools needed to solve the query that are in the list of top 'N' fetched tools.}
    \label{table: retriever-eval}
\end{table}

\clearpage

\section{The curious case of GitHub Copilot}
\begin{figure*}[hb]
    \centering
    \includegraphics[width = 0.9\textwidth]{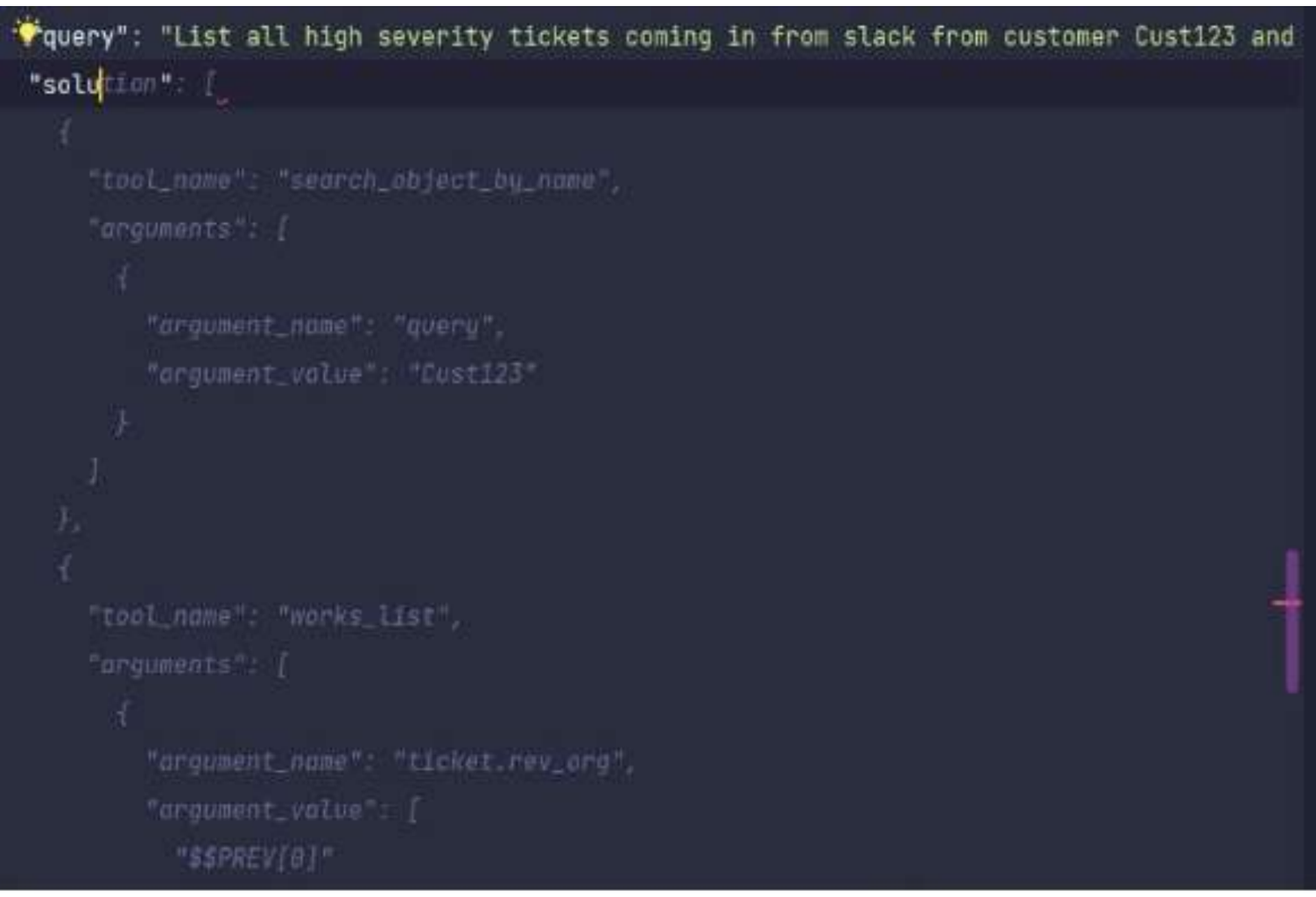}
    \caption{GitHub Copilot auto-completing the entire user query}
    \label{fig:Github Copilot}
\end{figure*}

We observed that Copilot takes a query and gives out a set of tools to complete the query. In fact, it was seen that the auto-complete system is capable of entirely auto-completing the user queries.
We observe that the official GitHub Copilot CLI banner shows a user input like- \emph{How do I find all the files bigger than ... }. This is very similar to our problem statement. This presents an exciting future opportunity to explore the methodology used when training GitHub Copilot. We show an example in Figure \ref{fig:Github Copilot}

\clearpage

\section{An example of our generated dataset}
\subsection{An Example Tool}

\label{appendix: stepback prompting}
\begin{listing}[!ht]
\begin{minted}{Text}
{
        "tool_name": "merge_work_items",
        "tool_description": "Combines multiple work items into a single item.",
        "args": [
            {
                "argument_name": "source_work_ids",
                "argument_type": "str",
                "is_array": true,
                "is_required": true,
                "argument_description": "The IDs of the source work items to merge.",
                "example": [
                    "TASK-123",
                    "ISSUE-456"
                ]
            },
            {
                "argument_name": "target_work_id",
                "argument_type": "str",
                "is_required": true,
                "argument_description": "The ID of the target work item to merge into.",
                "example": "TASK-789"
            }
        ],
        "output": {
            "argument_type": "object",
            "is_array": false,
            "is_required": true
        }
    }
\end{minted}
    \caption{An example of a tool generated by us}
    \label{listing: StepBack Prompting}
\end{listing}
\clearpage
\subsection{An Example datapoint in our generated dataset}
\label{appendix: example datapoint}
\begin{listing}[!ht]
\begin{minted}{Text}
{
    "query": "Summarize the works created by user DEVU-123 that are currently 'In Progress'",
    "solution": [
        {
            "tool_name": "works_list",
            "arguments": [
                {
                    "argument_name": "created_by",
                    "argument_value": [
                        "DEVU-123"
                    ]
                }
            ]
        },
        {
            "tool_name": "filter_by_status",
            "arguments": [
                {
                    "argument_name": "work_ids",
                    "argument_value": "$$PREV[0]"
                },
                {
                    "argument_name": "status",
                    "argument_value": [
                        "In Progress"
                    ]
                }
            ]
        },
        {
            "tool_name": "summarize_objects",
            "arguments": [
                {
                    "argument_name": "objects",
                    "argument_value": "$$PREV[1]"
                }
            ]
        }
    ]
}
\end{minted}
    \caption{An example of query json pair with retrieved generated by us on the tool dataset given in the problem statement and generated by us }
    \label{listing: StepBack Prompting}
\end{listing}
\clearpage

\section{StepBack Prompting}
\label{appendix: stepback prompting}
\begin{listing}[!ht]
\begin{minted}{Text}
Given the set of tools provided in the JSON format, here is an example scenario where we want to identify work items created by the current user that are high priority and add selected items to the current sprint:

Question: How can we find high priority work items created by the current user and add them to the current sprint?

Sub-Question 1: Which tool to select now to identify the current user?
Answer 1: We will use the tool "who_am_i" to get the id of the current user. The answer is "who_am_i".

Sub-Question 2: Which tool to select now to get the current sprint id?
Answer 2: After identifying the current user, we should use the tool "get_sprint_id" to get the ID of the current sprint. The answer is "get_sprint_id".

Sub-Question 3: Which tool to select now to list the work items created by the current user?
Answer 3: With the current user ID, we can utilize the "works_list" tool to retrieve a list of work items matching the request. We need to specify the "created_by" argument, which requires the user ID. The answer is "works_list".

Sub-Question 4: Which tool to select now to prioritize the listed work items?
Answer 4: Given the list of work items, we need to use the "prioritize_objects" tool to sort these items by priority. The answer is "prioritize_objects".

Sub-Question 5: Which tool to select now to summarize the high-priority work items?
Answer 5: The output from "prioritize_objects" should be fed into the "summarize_objects" tool to get a summary of the high-priority work items. The answer is "summarize_objects".

Sub-Question 6: Which tool to select now to add the high-priority work items to the current sprint?
Answer 6: We will use "add_work_items_to_sprint" to add the given high-priority work items to the sprint. This requires the work item IDs and the sprint ID. The answer is "add_work_items_to_sprint".

Sub-Question 7: Which tool to select now to validate that the work items are added?
Answer 7: No tool is needed to validate the addition of work items directly. Instead, we can use "works_list" again to confirm if the items are part of the sprint by using appropriate arguments such as "sprint_id" and "owned_by". The answer is "works_list".

Since we have completed the sequence of actions needed to reach a logical conclusion, we will add an end tool.

Sub-Question 8: Now we can answer the question: Which tool do we use to end the process?
Answer 8: The "end_tool" indicates that the process is complete. The answer is "end_tool".
\end{minted}
    \caption{StepBack Prompting with 100\% results using GPT-4}
    \label{listing: StepBack Prompting}
\end{listing}

\clearpage
\section{ControlLLM Prompting}
\label{appendix: ControlLLM Prompt}

\begin{listing}[!ht]
\begin{minted}{Text}
The following is a friendly conversation between a human and an AI. The AI is professional and parses user input to several tasks with lots of specific details from its context. If the AI does not know the answer to a question, it truthfully says it does not know. The AI assistant can parse user input to several tasks with JSON format as follows:<Solution> [“description”: task description, “tool_name”: tool name, “id”: task_id, “dep”: dependency_task_id, “arguments”: [“argument_name”: name of the argument the tool expects, “argument_value”: value of the specific argument or $$PREV[dependency_task_id]]]</Solution>. The ”description” should describe the task in detail, and AI assistant can add some details to improve the user’s request without changing the user’s original intention. The special tag “dependency_task_id” refers to the one in the dependency task (Please consider whether the dependency task generates arguments required by this tool.) and “dependency_task_id” must be in “dep” list. The “dep” list denotes the ids of the previous prerequisite tasks, which generate a new resource that the current task relies on. The special tag “task_id” must be integers, which refers to the order in which the tasks should be implemented.  The “arguments” field denotes the input resources this tool expects to execute the task. The “tool_name” MUST be selected from the following options: “works_list”, “summarize_objects”, “prioritize_objects”, “add_work_items_to_sprint”, “get_sprint_id”, “get_similar_work_items”, “search_object_by_name”, “create_actionable_tasks_from_text”, “who_am_i”, nothing else. Think step by step about all the tasks that can resolve the user’s request. Parse out as few tasks as possible while ensuring that the user request can be resolved. Pay attention to the dependencies and order among tasks. If some inputs of tools are not found, you cannot assume that they already exist. You can think a new task to generate those args that do not exist or ask for the user’s help. If the user request can’t be parsed, you need to reply empty JSON []. You should always respond in the following format: <Solution><YOUR_SOLUTION></Solution>.
<YOUR_SOLUTION> should be strict with JSON format described above.
Your knowledge base consists of tool descriptions and argument descriptions as explained below:
[
  {
    "tool_name": "who_am_i",
    "tool_description": "Returns the id of the current user",
    "args": [],
    "output": {
      "arg_type": "string",
      "is_array": false,
      "is_required": true
    }
  }
]
\end{minted}
    \caption{ControlLLM Prompting with 100\% results using GPT-4}
\end{listing}

\clearpage

\section{Task Decomposition Prompting}
\label{appendix: TASK DEC}

\begin{listing}[!ht]
\begin{minted}{Text}
You are a helpful assistant. Your job is to decompose a user query into tasks that can be solved with one tool each.
        Analyse the list of tools and split the query into tasks. Solve the problem by thinking step by step.
        In each thought, think about what the next step should be in order to solve the problem, based on the available tools. Explain why the tool is required.
        Find the required tool that should be called (make sure it is related to the thought), and construct a task to be completed using it, based on the thought. If no tool is required, use "no_tool".
        Ensure that each task contains the necessary information to call the tool associated with it. Do not create unnecessary steps, if they haven't been mentioned in the query. Keep the minimum number of required tools. Be short and concise.
        The output of the ith task can be referenced using "$$PREV[i]" (starts from 0). DO NOT call tools or write any code in the arguments, and do not make up arguments that don't exist. 
        
        
        If the query cannot be solved with the given tools, just return an empty list []
        
        Note that this is a product management system, and we call our customers revs. Objects are things like customers, parts, and users.
        {formatted_tools(retrieved_tools)}

        Example:
        Query: Summarize high severity tickets from the customer (rev) UltimateCustomer and add them to the current sprint.
        Solution:
        [
            {
                "thought": "First, we need to get the id of the customer UltimateCustomer",
                "tool_name": "search_object_by_name",
                "task": "Use the search_object_by_name tool with the argument 'query' = UltimateCustomer"
            },
            {
                "thought": "Next, we need to get the high severity tickets for the customer",
                "tool_name": "works_list",
                "task": "Use the works_list tool with the arguments: 'issue.rev_orgs' = "$$PREV[0]", 'ticket_severity' = 'high', and 'type' = 'ticket'."
            },
            {
                "thought": "Now, we need to summarize the high severity tickets obtained from the previous task",
                "tool_name": "summarize_objects",
                "task": "Use the summarize_objects tool with the output of the second task, argument 'objects' = "$$PREV[1]""
            },
            {
                "thought": "In order to add the tool to the current sprint, we need to get the current sprint ID",
                "tool_name": "get_sprint_id",
                "task": "Use the get_sprint_id tool to get the current sprint ID"
            },
            {
                "thought": "Finally, we need to add the work items obtained using works_list in the second task to the current sprint, whose ID was obtained in the previous task.",
                "tool_name": "add_work_items_to_sprint",
                "task": "Use the add_work_items_to_sprint tool with arguments 'work_ids'='$$PREV[1]' and 'sprint_id'='$$PREV[3]'
            }
        ]
\end{minted}
    \caption{Task Decomposition Prompt for OpenChat with retrieved tools}
\end{listing}

\clearpage
\section{Tool JSON Formation Prompting}
\label{appendix: Tool Formation}

\begin{listing}[!ht]
\begin{minted}{Text}
You are a helpful assistant. Your job is to output a json file which can be used to call the tool given below, based on the task given to you.
        These tasks are required in order to solve a given query. In case any arguments are missing in the task, you can still add them to the json.
        The output of the ith task can be referenced using "$$PREV[i]" (starts from 0). This is important, since many queries require a composition of tools.
        You can't reference tools that haven't been called yet.
        
        Tools must be explicitly called and cannot be called inside the arguments.
        

        Example:
        Query : Obtain work items from the customer support channel, summarize the ones related to part 'FEAT-345' part and prioritize them.
        
        Completed tasks and thought process:
        Task 0:
        Thought: First, retrieve all work items from the customer support channel related to 'FEAT-345'
        Tool_name: works_list
        Task: "Use the 'works_list' tool with the arguments 'ticket.source_channel'= ['customer support'] and 'applies_to_part'= ["FEAT-345"]
        
        Task 1:
        Thought: Next, summarize the work items related to 'FEAT-345' for clarity.
        Tool_name: summarize_objects
        Task: Use the 'summarize_objects' tool with the 'objects' argument being the output from the 'works_list' tool, 'objects'='$$PREV[0]'
        
        Your Task: 
        Task 2:
        Thought: Finally, prioritize the issues from the customer support channel that are urgent.
        Tool_name: "prioritize_objects
        Task: "Use the 'prioritize_objects' tool with the 'objects' argument being the output from the 'works_list' tool, argument 'objects' = '$$PREV[1]'
        
        Answer:
        {
            "tool_name": "prioritize_objects",
            "arguments": [
                {
                    "argument_name": "objects",
                    "argument_value": "$$PREV[1]"
                }
            ]
        }
\end{minted}
    \caption{Tool JSON Formation Prompt for OpenChat with retrieved tools}
\end{listing}

\begin{table}[htbp]
    \centering
    \begin{adjustbox}{width=\textwidth}
    \begin{tabular}{|c|p{9cm}|}
        \hline
        \textbf{Tool} & \textbf{Description} \\
        \hline
        \textbf{works\_list} & Returns a list of work items matching the request \\
        \hline
        \textbf{summarize\_objects} & Summarizes a list of objects. The logic of how to summarize a particular object type is an internal implementation detail \\
        \hline
        \textbf{prioritize\_objects} & Returns a list of objects sorted by priority. The logic of what constitutes priority for a given object is an internal implementation detail \\
        \hline
        \textbf{add\_work\_items\_to\_sprint} & Adds the given work items to the sprint \\
        \hline
        \textbf{get\_sprint\_id} & Returns the ID of the current sprint \\
        \hline
        \textbf{get\_similar\_work\_items} & Returns a list of work items that are similar to the given work item \\
        \hline
        \textbf{search\_object\_by\_name} & Given a search string, returns the id of a matching object in the system of record. If multiple matches are found, it returns the one where the confidence is highest. \\
        \hline
        \textbf{create\_actionable\_tasks\_from\_text} & Given a text, extracts actionable insights, and creates tasks for them, which are kind of a work item. \\
        \hline
        \textbf{who\_am\_i} & Returns the string ID of the current user \\
        \hline
    \end{tabular}
    \end{adjustbox}
    \caption{Tools Table} 
    \label{tab:my_table}
\end{table}


\end{appendices}

\end{document}